\documentclass[runningheads]{llncs}
\usepackage[T1]{fontenc}
\usepackage{subcaption}
\usepackage{amsmath}
\usepackage{amssymb}  
\usepackage{hyperref}
\usepackage{makecell}
\usepackage{adjustbox}
\usepackage{multirow}
\usepackage{enumitem}
\usepackage{wasysym}
\usepackage{xcolor}
\usepackage{booktabs, pifont}
\newcommand{\cmark}{\textcolor{green!60!black}{\ding{51}}}%
\newcommand{\xmark}{\textcolor{red!70!black}{\ding{55}}}%
\usepackage{fontawesome5}

\usepackage{graphicx,verbatim}
\begin{document}
\title{Longitudinal Multi-View Modeling for Breast Cancer Risk Prediction}

\titlerunning{Longitudinal Multi-View Breast Cancer Risk Prediction}
\author{Solveig Thrun\inst{1} \and
Zijun Sun\inst{1}  \and   Suaiba A. Salahuddin\inst{1} \and
  Kristoffer Wickstrøm\inst{1} \and Elisabeth Wetzer\inst{1} \and Stine Hansen\inst{2} \and Robert Jenssen\inst{1,3,4}\ \and Michael Kampffmeyer\inst{1,3}
}
\authorrunning{S. Thrun et al.}

%
\institute{Department of Physics and
Technology, UiT The Arctic University of Norway, Tromsø, Norway 
\email{solveig.thrun@uit.no}\\ \and SPKI The Norwegian Centre for Clinical Artificial Intelligence, University Hospital of North Norway, Tromsø, Norway \\ \and
Norwegian Computing Center, Oslo, Norway \and
Pioneer Centre for AI, University of Copenhagen, Copenhagen, Denmark }
 
 \maketitle              
\begin{abstract} 
Accurate breast cancer risk prediction from screening mammography is critical for enabling personalized screening intervals and early detection. Recent deep learning methods have shown the value of longitudinal data and explicit temporal alignment. However, existing approaches either perform explicit alignment using a single mammographic view or model multiple views without explicit longitudinal alignment, limiting their ability to exploit the complementary spatial–temporal information used in clinical practice.
To address this gap, we propose LMV-Net, a longitudinal multi-view breast cancer risk prediction model that jointly analyzes anatomically complementary CC and MLO views within an explicitly aligned longitudinal framework. We evaluate our approach on the public EMBED and CSAW-CC datasets, comparing it to state-of-the-art breast cancer risk prediction methods. Our model consistently outperforms existing approaches in overall risk prediction performance and across different breast density and cancer subgroups. Importantly, these improvements highlight the potential of longitudinal multi-view modeling to enhance risk stratification, paving the way for future work on personalized screening, earlier identification of high-risk patients, and more efficient screening resource allocation. The code is available at  \url{https://github.com/sot176/LMV-Net}.

\end{abstract}
\section{Introduction}
Breast cancer is the most common cancer among women worldwide, representing a major public health challenge~\cite{cancerstat}. Early detection through population-based screening has been critical in reducing mortality, with mammography, typically acquired in craniocaudal (CC) and mediolateral oblique (MLO) views, remaining the gold standard~\cite{pharmaceutics13050723,zielonke2020evidence}. Large-scale screening programs have therefore played a key role in identifying cancers at an earlier, better treatable stage~\cite{zielonke2020evidence}.
However, effective risk stratification for personalized screening and early detection in high-risk populations remains limited~\cite{rubio2024risk}, especially in women with dense breast tissue, where density both increases cancer risk and reduces mammographic sensitivity~\cite{bodewes2022density,cancerriskdensity}.

Recent advances in deep learning have improved breast cancer risk prediction from mammography, offering a data-driven path toward more individualized screening~\cite{2024riskreview}. Early deep learning models analyzed current images in isolation (e.g., Mirai~\cite{mirai}), overlooking the valuable longitudinal information that radiologists exploit when comparing current and prior examinations~\cite{akwo2024access,prioirmammo}. To leverage this longitudinal information, recent models incorporate prior screenings, enabling them to capture gradual tissue changes over time and improve risk assessment~\cite{dadsetan2022deep,changesrisk,Kar_Longitudinal_MICCAI2024,miccai2023,vmramar,Wan_Ordinal_MICCAI2024}. These models consistently outperform single-time-point approaches, demonstrating the value of modeling temporal information.

Building on advances in mammography research, most longitudinal mammography models rely on implicit alignment (e.g., LoMaR~\cite{Kar_Longitudinal_MICCAI2024}, VMRA-MaR~\cite{vmramar}), comparing prior and current images without enforcing spatial correspondence. Although non-rigid breast deformation makes explicit alignment challenging, recent evidence indicates that such strategies are better suited to capture subtle temporal variations, outperforming implicit approaches for modeling longitudinal change~\cite{thrunarxiv2025}. Nevertheless, existing explicit alignment methods (ImgFeatAlign~\cite{thrunarxiv2025,mammoregnet}, OA-BreaCR~\cite{Wan_Ordinal_MICCAI2024}) have been restricted to single-view formulations, without jointly modeling the complementary information available across CC and MLO views. As a result, current approaches do not fully exploit the combined temporal and cross-view cues that are critical for accurate breast cancer risk prediction.

To address this shortcoming, we introduce a breast-wise longitudinal multi-view risk prediction framework that explicitly aligns the feature maps of each CC and MLO view across multiple time points, capturing subtle longitudinal changes and more closely mirroring radiologists’ workflow. Our approach uses image-based deformation fields to align current and prior feature maps for each view, mitigating spatial misalignment while preserving discriminative information. By introducing a dual-stream attention mechanism, the model enables multi-view, breast-wise comparison over time. This design effectively captures longitudinal tissue changes and provides a foundation for more reliable risk prediction. Ultimately, it supports earlier identification of high-risk patients and facilitates personalized screening strategies.

Our \textbf{contributions} are:
\begin{enumerate}[noitemsep,  topsep=-1.2pt]     
\item LMV-Net, a breast-wise longitudinal multi-view model for breast cancer risk prediction, that facilitates multi-view, multi-time-point integration via explicit feature-space alignment and cross-view temporal fusion.
\item A Dual Stream Attention block that refines longitudinal features of each view (CC and MLO) using parallel self- and cross-attention.
\item Extensive validation on two large public datasets, EMBED and CSAW-CC, demonstrating robust performance.
\end{enumerate}

\section{Method}
\begin{figure}[tb]
    \centering
    \includegraphics[width=1.0\textwidth]{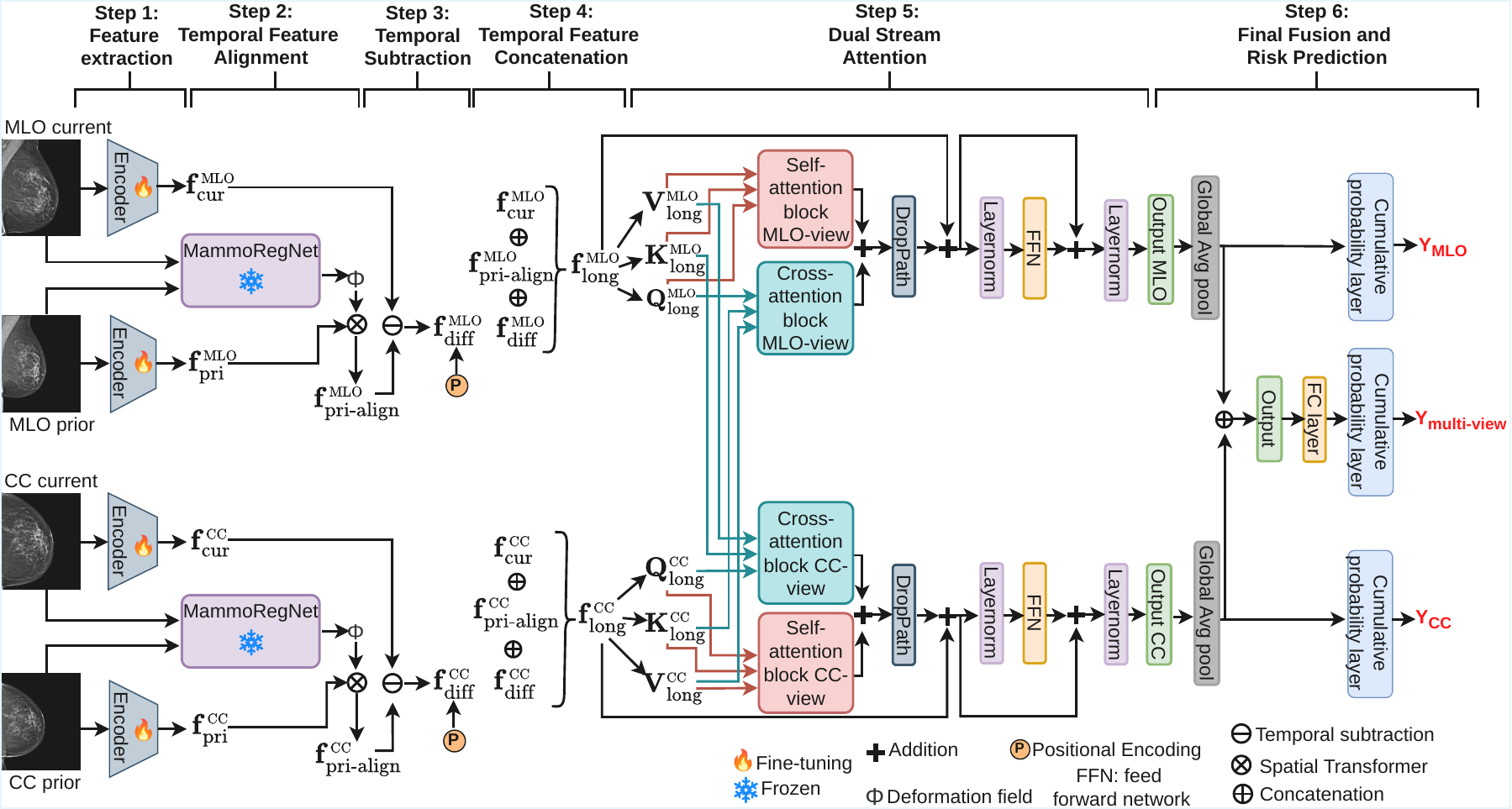}
    \caption{Schematic overview of the proposed LMV-Net. }
    \label{fig:lmv-model}
\end{figure}

\noindent We aim to predict an individual’s future breast cancer risk over a five-year horizon using mammograms from current and prior screening visits for a single breast, to support personalized, risk-based screening programs. Unlike prior approaches that rely on explicit alignment and use only a single view~\cite{thrunarxiv2025,Wan_Ordinal_MICCAI2024}, our proposed LMV-Net (Fig.~\ref{fig:lmv-model}) incorporates both CC and MLO views, which provide complementary information. To achieve this, we introduce a dual-stream attention block for multi-view fusion that refines explicitly aligned longitudinal features through parallel self- and cross-attention. LMV-Net comprises six sequential stages. \\
\textbf{Step 1: Feature extraction.} A shared image encoder extracts feature maps from current and prior mammograms in the MLO and CC views, yielding four feature maps:
current feature maps \(\mathbf{f}_{\text{cur}}^{\scriptscriptstyle \mathrm{MLO}}, \mathbf{f}_{\text{cur}}^{\scriptscriptstyle \mathrm{CC}} \in \mathbb{R}^{C \times H \times W}\) and prior feature maps
\(\mathbf{f}_{\text{pri}}^{\scriptscriptstyle \mathrm{MLO}}, \mathbf{f}_{\text{pri}}^{\scriptscriptstyle \mathrm{CC}} \in \mathbb{R}^{C \times H \times W}\), where \(C\) is the number of channels, and \(H\) and \(W\) denote the height and width of the feature maps, respectively. \\
\textbf{Step 2: Temporal feature alignment.}
Inspired by~\cite{thrunarxiv2025}, we explicitly align longitudinal feature maps in feature space using image-based deformation fields. The deformation field is obtained using a pretrained and frozen MammoRegNet~\cite{mammoregnet} by registering the prior mammogram to the current mammogram. A separate deformation field is estimated for each view (CC and MLO). Using a spatial transformer function~\cite{mammoregnet,Wan_Ordinal_MICCAI2024}, the prior feature maps (\(\mathbf{f}_{\text{pri}}^{\scriptscriptstyle \mathrm{MLO}}\) and \(\mathbf{f}_{\text{pri}}^{\scriptscriptstyle \mathrm{CC}}\)) are aligned to the corresponding current feature maps (\(\mathbf{f}_{\text{cur}}^{\scriptscriptstyle \mathrm{MLO}}\) and \(\mathbf{f}_{\text{cur}}^{\scriptscriptstyle \mathrm{CC}}\)), resulting in the aligned prior feature maps \(\mathbf{f}_{\text{pri-align}}^{\scriptscriptstyle \mathrm{MLO}}\) and \(\mathbf{f}_{\text{pri-align}}^{\scriptscriptstyle \mathrm{CC}}\). \\
\textbf{Step 3: Temporal subtraction.}   
To capture temporal changes, we follow~\cite{thrunarxiv2025,Wan_Ordinal_MICCAI2024} and compute view-specific difference features via temporal subtraction:
\(\mathbf{f}_{\text{diff}}^{\scriptscriptstyle \mathrm{CC}} = \mathbf{f}_{\text{cur}}^{\scriptscriptstyle \mathrm{CC}} - \mathbf{f}_{\text{pri-align}}^{\scriptscriptstyle \mathrm{CC}},\quad
\mathbf{f}_{\text{diff}}^{\scriptscriptstyle \mathrm{MLO}} = \mathbf{f}_{\text{cur}}^{\scriptscriptstyle \mathrm{MLO}} - \mathbf{f}_{\text{pri-align}}^{\scriptscriptstyle \mathrm{MLO}}\). These difference features encode longitudinal changes between aligned current and prior mammograms, and positional encoding~\cite{Wan_Ordinal_MICCAI2024} is applied to \(\mathbf{f}_{\text{diff}}^{\scriptscriptstyle \mathrm{CC}}\) and \(\mathbf{f}_{\text{diff}}^{\scriptscriptstyle \mathrm{MLO}}\) to account for variable time intervals between screenings.
 \\
\textbf{Step 4: Temporal feature concatenation.} Current, aligned prior, and difference features are concatenated along the channel dimension to form $\mathbf{f}_{\text{long}}^{\scriptscriptstyle \mathrm{MLO}}\in \mathbb{R}^{3C \times H \times W}$ and $ \mathbf{f}_{\text{long}}^{\scriptscriptstyle \mathrm{CC}} \in \mathbb{R}^{3C \times H \times W}$, integrating current appearance, longitudinal context, and temporal changes. \\
\textbf{Step 5: Dual Stream Attention.} We propose a Dual Stream Attention block designed to enhance the modeling of longitudinal multi-view features. The module jointly refines the longitudinal representations \(\mathbf{f}_{\text{long}}^{\scriptscriptstyle \mathrm{MLO}}\) and \(\mathbf{f}_{\text{long}}^{\scriptscriptstyle \mathrm{CC}}\) by explicitly capturing both intra-view and inter-view dependencies. For each view, the block applies multi-head self-attention to model long-range intra-view interactions and improve spatial coherence within the same mammographic view. In parallel, multi-head cross-attention leverages features from the complementary view as keys and values, enabling each view to selectively attend to anatomically and semantically relevant regions in the other view. This mechanism facilitates effective information exchange between CC and MLO representations, enabling the model to leverage complementary perspectives of the same breast tissue.
The outputs of self- and cross-attention are fused via element-wise summation, after which stochastic depth (DropPath~\cite{droppath}) is applied to improve generalization, and reduce overfitting. This attention output is added to the original residual features, followed by layer normalization~\cite{layernorm}. Each view is subsequently processed through a feed-forward network (FFN) with a fourfold expansion of the channel dimension, increasing model expressiveness. A second residual connection and normalization are applied after the FFN. Through this dual-stream attention design, each view integrates both its own global spatial context and complementary cross-view information, resulting in enriched and more discriminative multi-view feature representations.  \\
\textbf{Step 6: Final fusion and risk prediction.}
The refined \(\mathbf{f}_{\text{long}}^{\scriptscriptstyle \mathrm{MLO}}\) and \(\mathbf{f}_{\text{long}}^{\scriptscriptstyle \mathrm{CC}}\) features are pooled and concatenated into a fused multi-view representation $\mathbf{f}$, which passes through a cumulative probability layer~\cite{miccai2023,mirai} for time-dependent risk estimation. This layer comprises two fully connected layers: one predicting hazards at each future time step (i.e. 1 year, 2 years, …) and another computing a baseline hazard, with ReLU enforcing non-negativity. The cumulative probability at time $t$ is then computed as
\begin{equation}
P(t \mid \mathbf{f}) = h_{\text{base}}(\mathbf{f}) + \sum_{j=1}^t h_j(\mathbf{f}),
\end{equation}
where $h_{\text{base}}(\mathbf{f})$ denotes the baseline hazard and $h_j(\mathbf{f})$ the hazard at time step $j$, capturing the sequential accumulation of risk.
Training uses a time-dependent binary cross-entropy (BCE) loss, with labels set to 1 if cancer occurs within $t$ years and 0 otherwise. For censored patients, a masking function $\delta(t)$~\cite{Wan_Ordinal_MICCAI2024,mirai} is applied so that only valid time points contribute to the summed BCE loss,. Following~\cite{mvpnet,Wan_Ordinal_MICCAI2024}, we adopt multilevel joint learning to capture complementary dependencies between CC and MLO views. The network jointly learns a fused multi-view representation for risk estimation and view-specific features that serve as auxiliary supervision during training, encouraging discriminative per-view learning and enhancing the quality of the fused representation used at inference.

\section{Experimental Setup and Results}

\begin{table}[tb]
\centering
\scriptsize
\caption{
Breast-level CC+MLO exam splits and time-to-cancer labels.
}
\label{tab:timecancerdataset}

\begin{subtable}[t]{0.48\linewidth}
\centering
\begin{tabular}{|c|l|cccccc|c|}
\hline
\multirow{5}{*}{\rotatebox{90}{\textbf{EMBED}}}
 &  & \multicolumn{6}{c|}{\textbf{Positive}} & \textbf{Negative} \\
\cline{3-8} \cline{9-9} &
\textbf{Split} & \textbf{1Y} & \textbf{2Y} & \textbf{3Y} & \textbf{4Y} & \textbf{5Y} & \textbf{All} & \textbf{$>$5Y} \\
\cline{2-9}
&Train & 370 & 96 & 83 & 46 & 34 & 629 & 12262 \\
 &Val   & 146 & 32 & 29 & 20 & 14 & 241 & 4907  \\
&Test  & 220 & 76 & 54 & 37 & 29 & 416 & 7341  \\
\hline
\end{tabular}
\end{subtable}
\hfill
\begin{subtable}[t]{0.48\linewidth}
\centering
\begin{tabular}{|c|l|cccccc|c|}
\hline
\multirow{5}{*}{\rotatebox{90}{\makecell{\textbf{CSAW-}\\\textbf{CC}}}}
 &  & \multicolumn{6}{c|}{\textbf{Positive}} & \textbf{Negative} \\
\cline{3-8} \cline{9-9} &
\textbf{Split} & \textbf{1Y} & \textbf{2Y} & \textbf{3Y} & \textbf{4Y} & \textbf{5Y} & \textbf{All} & \textbf{$>$5Y} \\
\cline{2-9}
&Train & 352 & 350 & 94 & 224 & 88 & 1108 & 15686 \\
&Val   & 140 & 142 & 48 & 92 & 28 & 450 & 6188  \\
&Test  & 208 & 214 & 76 & 124 & 54 & 676 & 9402  \\
\hline
\end{tabular}
\end{subtable}

\end{table}

\textbf{Datasets and Pre-processing}
We evaluate LMV-Net on two large, publicly available mammography datasets: the Emory Breast Imaging Dataset (EMBED)~\cite{embed}
 and the Cohort of Screen-Aged Women Case Control dataset (CSAW-CC)~\cite{csawcc}, ensuring reproducibility and an experimental setup mirroring prior work~\cite{Kar_Longitudinal_MICCAI2024,vmramar,thrunarxiv2025,mammoregnet,Wan_Ordinal_MICCAI2024,mirai}. EMBED comprises mammograms from a diverse population acquired on Hologic, GE, and Fujifilm systems, while CSAW-CC contains cases collected at Karolinska University Hospital (Sweden) using Hologic systems. Following~\cite{Wan_Ordinal_MICCAI2024}, we include only patients with at least five years of follow-up. All standard 2D mammograms are resized to $1664 \times 2048$ pixels while preserving aspect ratio. Data are split at the patient level into training, validation, and test sets (5:2:3). CSAW-CC provides LIBRA-based breast density estimates~\cite{libra}, grouped into three equally sized categories (low, medium, high), whereas EMBED uses BI-RADS density categories~\cite{birads}. Time-to-cancer is defined as the interval between image acquisition and diagnosis. Table~\ref{tab:timecancerdataset} summarizes dataset splits and time-to-cancer distributions at the breast-wise CC–MLO exam level. For patients with more than two time points, we construct all consecutive time-point pairs of images for training and evaluation.
For training MammoRegNet, we exclusively used patients who were excluded from the risk prediction cohort, preventing any potential data leakage.

\noindent \textbf{Evaluation Metrics} 
Following~\cite{miccai2023,Wan_Ordinal_MICCAI2024,mirai}, we evaluate performance using the C-index~\cite{c_index} and AUC~\cite{auc} over 1–5 year risk horizons. 95\% confidence intervals (CIs) are estimated via exam-level bootstrapping with 1,000 resamples. Statistical significant improvements over the second-best model is assessed using a paired non-parametric bootstrap test, with $p<0.05$ considered significant.\\
\textbf{Implementation Details} 
All models were implemented in PyTorch 2.0.1~\cite{pytorch} and on eight AMD Instinct MI210 GPUs (64 GB each) with a batch size of 4 per GPU (32 total). For fair comparison, we use the Mirai encoder~\cite{mirai}, a ResNet-18~\cite{resnet18} with the final fully connected layer removed, following prior work~\cite{Kar_Longitudinal_MICCAI2024,vmramar,thrunarxiv2025,Wan_Ordinal_MICCAI2024}. Self- and cross-attention use four heads, and training uses AdamW~\cite{adamw} with a learning rate of $5 \times 10^{-5}$ and weight decay of $1 \times 10^{-4}$. 
LMV-Net was trained for up to 30 epochs, with early stopping after 15 stagnant epochs based on the validation C-index and learning rate scheduling that halved the learning rate after 5 stagnant epochs. Data augmentation included random cropping, affine transformations, color jitter, and gamma correction. Baselines~\cite{vmramar,mammoregnet,Wan_Ordinal_MICCAI2024,mirai} and MammoRegNet~\cite{thrunarxiv2025,mammoregnet} followed official implementations and hyperparameters, and LMV-Net was evaluated with both frozen and fine-tuned pretrained Mirai encoders.

\subsection{Results}

\begin{table}[tb]
\centering
\scriptsize
\caption{1–5 Year breast cancer risk prediction: C-index and AUC values with $\pm$ 95\% CI. \textcolor{blue}{\scalebox{0.9}{\faSnowflake}}: frozen encoder,  
\textcolor{orange}{\scalebox{0.9}{\faFire}}: fine-tuned encoder. $\Delta$ denotes \% improvement of best LMV-Net over best baseline. $^{*}$ indicates statistically significant improvement of LMV-Net over the best baseline (p < 0.05). Training time in hours (h) and total number of model parameters (\#, in millions).}
\begin{adjustbox}{center,max width=0.91\textwidth}
\begin{tabular}{|c|c|c|*{5}{c}|c|c|}
\hline
& \multirow{2}{*}{} & \multirow{2}{*}{C-index (\%) $\uparrow$} & \multicolumn{5}{c|}{Follow-up year AUC (\%) $\uparrow$} & \multirow{2}{*}{h} & \multirow{2}{*}{ \#} \\
\cline{4-8}
& & & 1-Y & 2-Y & 3-Y & 4-Y & 5-Y  & & \\
\hline

\multirow{7}{*}{\rotatebox{90}{\textbf{EMBED}}}
  
    & \makecell{Mirai~\cite{mirai} (\textcolor{blue}{\scalebox{0.9}{\faSnowflake}})}
& \makecell{59.8\fontsize{6}{6}\selectfont$\pm$8.7} 
& \makecell{56.7\fontsize{6}{6}\selectfont$\pm$20.7} 
& \makecell{52.6\fontsize{6}{6}\selectfont$\pm$15.3} 
& \makecell{54.0\fontsize{6}{6}\selectfont$\pm$12.6} 
& \makecell{49.3\fontsize{6}{6}\selectfont$\pm$12.5}
& \makecell{49.2\fontsize{6}{6}\selectfont$\pm$11.6} & 6& 28\\

    & \makecell{Mirai~\cite{mirai} (\textcolor{orange}{\scalebox{0.9}{\faFire}})} 
& \makecell{69.1\fontsize{6}{6}\selectfont$\pm$10.6} 
& \makecell{71.0\fontsize{6}{6}\selectfont$\pm$15.6} 
& \makecell{70.9\fontsize{6}{6}\selectfont$\pm$15.5}
& \makecell{70.6\fontsize{6}{6}\selectfont$\pm$11.7} 
& \makecell{65.7\fontsize{6}{6}\selectfont$\pm$11.3} 
& \makecell{65.9\fontsize{6}{6}\selectfont$\pm$11.6} & 7 & 28\\

    & \makecell{VMRA-MaR~\cite{vmramar} (\textcolor{orange}{\scalebox{0.9}{\faFire}})} 
& \makecell{72.4\fontsize{6}{6}\selectfont$\pm$12.5} 
& \makecell{77.9\fontsize{6}{6}\selectfont$\pm$4.5} 
& \makecell{75.3\fontsize{6}{6}\selectfont$\pm$9.7}
& \makecell{73.1\fontsize{6}{6}\selectfont$\pm$12.7}
& \makecell{68.9\fontsize{6}{6}\selectfont$\pm$14.3} 
& \makecell{67.1\fontsize{6}{6}\selectfont$\pm$11.8} & 45 & 2802\\

  & \makecell{OA-BreaCR~\cite{Wan_Ordinal_MICCAI2024} (\textcolor{orange}{\scalebox{0.9}{\faFire}})} 
  & \makecell{72.3\fontsize{6}{6}\selectfont$\pm$2.0}
  & \makecell{72.9\fontsize{6}{6}\selectfont$\pm$2.8}
  & \makecell{72.3\fontsize{6}{6}\selectfont$\pm$4.9}
  & \makecell{72.4\fontsize{6}{6}\selectfont$\pm$2.5}
  & \makecell{72.3\fontsize{6}{6}\selectfont$\pm$2.4} 
                      & \makecell{73.2\fontsize{6}{6}\selectfont$\pm$2.4} & 9& 15\\

  & \makecell{ImgFeatAlign~\cite{thrunarxiv2025} (\textcolor{blue}{\scalebox{0.9}{\faSnowflake}})} 
  & \makecell{74.7\fontsize{6}{6}\selectfont$\pm$2.4}
  & \makecell{75.0\fontsize{6}{6}\selectfont$\pm$2.9} 
  & \makecell{75.5\fontsize{6}{6}\selectfont$\pm$2.4} 
  & \makecell{75.3\fontsize{6}{6}\selectfont$\pm$2.1} 
  & \makecell{75.9\fontsize{6}{6}\selectfont$\pm$2.3} 
  & \makecell{72.5\fontsize{6}{6}\selectfont$\pm$3.2} &12 & 52 \\

  & \makecell{LMV-Net (\textcolor{blue}{\scalebox{0.9}{\faSnowflake}})}
  & \makecell{78.3$^{*}$\fontsize{6}{6}\selectfont$\pm$3.0}
  & \makecell{80.8$^{*}$\fontsize{6}{6}\selectfont$\pm$3.7}
  & \makecell{78.3$^{*}$\fontsize{6}{6}\selectfont$\pm$3.5} 
  & \makecell{77.5$^{*}$\fontsize{6}{6}\selectfont$\pm$3.3} 
  & \makecell{76.6\fontsize{6}{6}\selectfont$\pm$3.2} 
  & \makecell{75.1$^{*}$\fontsize{6}{6}\selectfont$\pm$4.9} & 15 & 65\\

 & \makecell{LMV-Net (\textcolor{orange}{\scalebox{0.9}{\faFire}})} 
& \makecell{\textbf{81.4$^{*}$}\textbf{\fontsize{6}{6}\selectfont $\pm$2.8}}
& \makecell{\textbf{83.9$^{*}$}\textbf{\fontsize{6}{6}\selectfont $\pm$3.4}}
& \makecell{\textbf{82.1$^{*}$}\textbf{\fontsize{6}{6}\selectfont $\pm$3.0}} 
& \makecell{\textbf{80.8$^{*}$}\textbf{\fontsize{6}{6}\selectfont $\pm$2.9}} 
& \makecell{\textbf{80.0$^{*}$}\textbf{\fontsize{6}{6}\selectfont $\pm$2.8}} 
& \makecell{\textbf{77.2$^{*}$}\textbf{\fontsize{6}{6}\selectfont $\pm$4.7}}  & 16 & 65\\
\hline
& \makecell{ $\Delta$ (\%) $\uparrow$}
& \makecell{\textbf{\textcolor{green!60!black}{+ 6.7}} }
& \makecell{\textbf{\textcolor{green!60!black}{+ 6.0}} }
& \makecell{\textbf{\textcolor{green!60!black}{+ 6.6}}} 
& \makecell{\textbf{\textcolor{green!60!black}{+ 5.5}}} 
& \makecell{\textbf{\textcolor{green!60!black}{+ 4.1}}} 
& \makecell{\textbf{\textcolor{green!60!black}{+ 4.0}}} & & \\
\Xhline{1.4pt}

\multirow{7}{*}{\rotatebox{90}{\textbf{CSAW-CC}}}

& \makecell{Mirai~\cite{mirai}(\textcolor{blue}{\scalebox{0.9}{\faSnowflake}})}
& \makecell{71.4\fontsize{6}{6}\selectfont$\pm$2.7} 
& \makecell{56.8\fontsize{6}{6}\selectfont$\pm$10.6} 
& \makecell{54.7\fontsize{6}{6}\selectfont$\pm$5.8} 
& \makecell{56.4\fontsize{6}{6}\selectfont$\pm$5.1} 
& \makecell{56.1\fontsize{6}{6}\selectfont$\pm$4.2} 
& \makecell{55.4\fontsize{6}{6}\selectfont$\pm$4.0} & 7 & 28\\

    & \makecell{Mirai~\cite{mirai} (\textcolor{orange}{\scalebox{0.9}{\faFire}})} 
& \makecell{72.3\fontsize{6}{6}\selectfont$\pm$2.9} 
& \makecell{69.9\fontsize{6}{6}\selectfont$\pm$3.3} 
& \makecell{69.8\fontsize{6}{6}\selectfont$\pm$3.9}
& \makecell{69.2\fontsize{6}{6}\selectfont$\pm$3.8} 
& \makecell{67.0\fontsize{6}{6}\selectfont$\pm$3.4} 
& \makecell{66.9\fontsize{6}{6}\selectfont$\pm$3.4} &8 & 28\\

    & \makecell{VMRA-MaR~\cite{vmramar} (\textcolor{orange}{\scalebox{0.9}{\faFire}})} 
& \makecell{72.7\fontsize{6}{6}\selectfont$\pm$2.6} 
& \makecell{70.9\fontsize{6}{6}\selectfont$\pm$5.5}
& \makecell{70.3\fontsize{6}{6}\selectfont$\pm$5.6} 
& \makecell{69.5\fontsize{6}{6}\selectfont$\pm$4.4} 
& \makecell{70.9\fontsize{6}{6}\selectfont$\pm$4.3} 
& \makecell{64.5\fontsize{6}{6}\selectfont$\pm$6.0} & 46& 2802\\

  & \makecell{OA-BreaCR~\cite{Wan_Ordinal_MICCAI2024} (\textcolor{orange}{\scalebox{0.9}{\faFire}})} 
& \makecell{61.6\fontsize{6}{6}\selectfont$\pm$4.2} 
& \makecell{63.5\fontsize{6}{6}\selectfont$\pm$4.4} 
& \makecell{61.7\fontsize{6}{6}\selectfont$\pm$2.1} 
& \makecell{65.0\fontsize{6}{6}\selectfont$\pm$2.0} 
& \makecell{67.0\fontsize{6}{6}\selectfont$\pm$1.9} 
& \makecell{67.7\fontsize{6}{6}\selectfont$\pm$2.2} & 10& 15\\

 & \makecell{ImgFeatAlign~\cite{thrunarxiv2025} (\textcolor{blue}{\scalebox{0.9}{\faSnowflake}})} 
& \makecell{70.4\fontsize{6}{6}\selectfont$\pm$1.9} 
& \makecell{72.0\fontsize{6}{6}\selectfont$\pm$2.4} 
& \makecell{71.5\fontsize{6}{6}\selectfont$\pm$1.9} 
& \makecell{72.6\fontsize{6}{6}\selectfont$\pm$1.9}
& \makecell{72.0\fontsize{6}{6}\selectfont$\pm$2.1} 
& \makecell{75.2\fontsize{6}{6}\selectfont$\pm$2.3} &13 & 52\\

& \makecell{LMV-Net (\textcolor{blue}{\scalebox{0.9}{\faSnowflake}})}
& \makecell{73.4$^{*}$\fontsize{6}{6}\selectfont$\pm$2.5}  
& \makecell{77.5$^{*}$\fontsize{6}{6}\selectfont$\pm$3.3}  
& \makecell{73.5$^{*}$\fontsize{6}{6}\selectfont$\pm$2.6}
& \makecell{74.0\fontsize{6}{6}\selectfont$\pm$2.5}
& \makecell{73.1\fontsize{6}{6}\selectfont$\pm$2.6} 
& \makecell{\textbf{75.9}\textbf{\fontsize{6}{6}\selectfont$\pm$2.8}}
& 16 & 65\\

 & \makecell{LMV-Net (\textcolor{orange}{\scalebox{0.9}{\faFire}})} 
& \makecell{ \textbf{74.4$^{*}$}\textbf{\fontsize{6}{6}\selectfont$\pm$2.7}} 
& \makecell{ \textbf{78.7$^{*}$}\textbf{\fontsize{6}{6}\selectfont$\pm$3.6} }
& \makecell{\textbf{74.1$^{*}$}\textbf{\fontsize{6}{6}\selectfont$\pm$2.9} }
& \makecell{\textbf{74.3$^{*}$}\textbf{\fontsize{6}{6}\selectfont$\pm$2.7}} 
& \makecell{ \textbf{73.5}\textbf{\fontsize{6}{6}\selectfont$\pm$2.8}} 
& \makecell{75.0\fontsize{6}{6}\selectfont$\pm$3.1} & 17& 65\\
\hline
& \makecell{ $\Delta$ (\%) $\uparrow$}
& \makecell{\textbf{\textcolor{green!60!black}{+ 1.7}} }
& \makecell{\textbf{\textcolor{green!60!black}{+ 6.7}}} 
& \makecell{\textbf{\textcolor{green!60!black}{+ 2.6}} }
& \makecell{\textbf{\textcolor{green!60!black}{+ 1.7}}} 
& \makecell{\textbf{\textcolor{green!60!black}{+ 1.5}}} 
& \makecell{\textbf{\textcolor{green!60!black}{+ 0.7}}} & &  \\
\hline

\end{tabular}

\end{adjustbox}
\label{tab:riskpred}
\end{table}

\noindent \textbf{Comparison with State-of-the-Arts (SOTA)}
Table~\ref{tab:riskpred} compares the proposed LMV-Net with SOTA methods, including Mirai~\cite{mirai} (Sci. Transl. Med.’2021), OA-BreaCR~\cite{Wan_Ordinal_MICCAI2024} (MICCAI’24), VMRA-MaR~\cite{vmramar} (MICCAI’25), and ImgFeatAlign~\cite{thrunarxiv2025} (MICCAI’25), on both datasets. LMV-Net consistently outperforms all baselines in C-index and AUC across all follow-up years on both datasets. Fine-tuning the encoder yields further improvements, achieving statistically significant gains ($p<0.05$) over the second-best model. These results highlight the benefit of jointly modeling multi-view and longitudinal information for breast cancer risk prediction.
 Notably, the larger gains observed on EMBED are due to its greater heterogeneity across populations and imaging systems, whereas the single-scanner CSAW-CC cohort is more homogeneous, limiting the potential benefits of improved model generalization.

\begin{figure}[tb]
    \centering

    \begin{minipage}[t]{0.465\textwidth}
        \centering
\includegraphics[width=\textwidth]{images/C-index_density_combined_new.png}
    \end{minipage}
    \hfill
    \begin{minipage}[t]{0.162\textwidth}
        \centering
        \includegraphics[width=\textwidth]{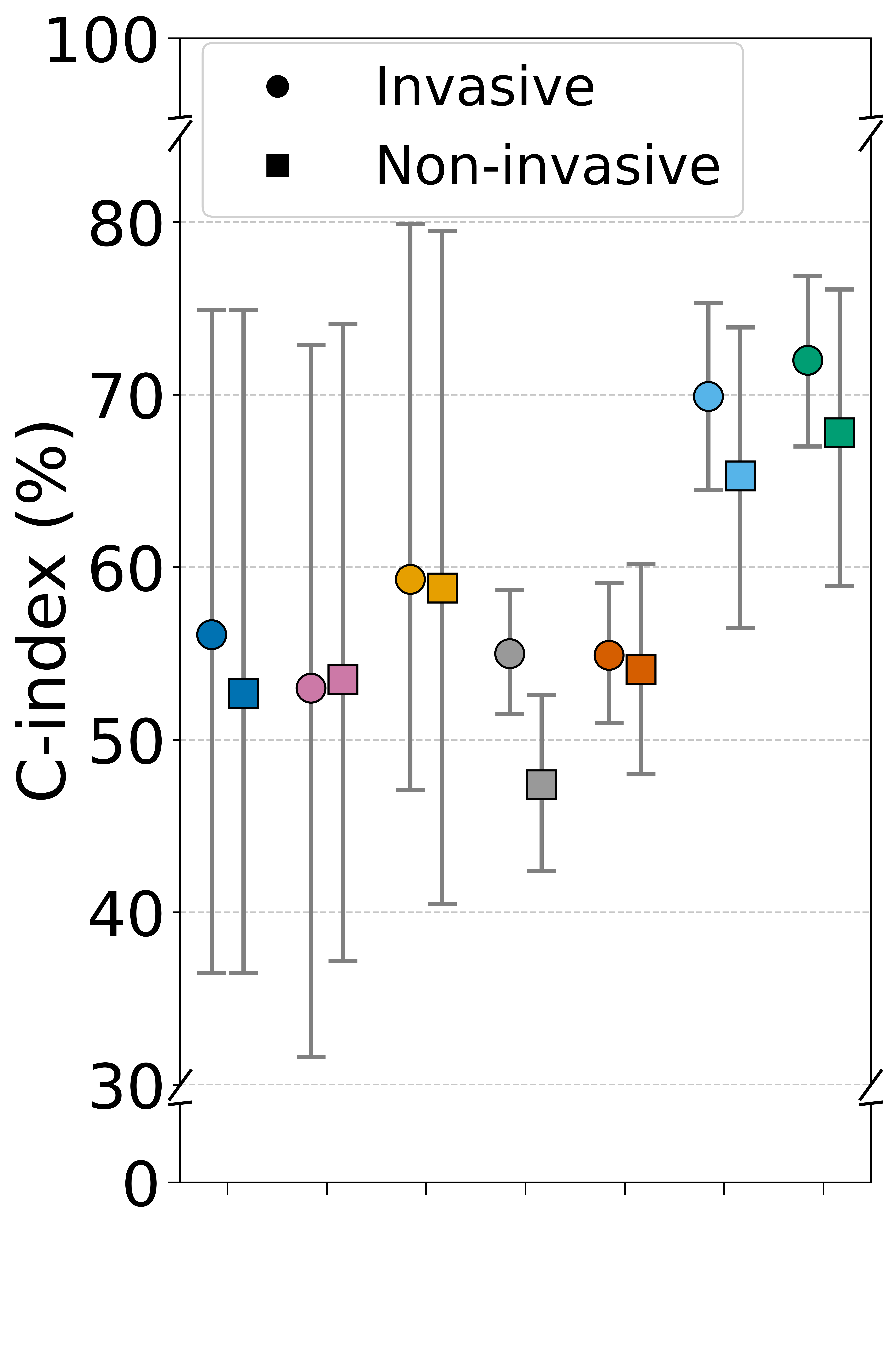}
    \end{minipage}
    \hfill
    \begin{minipage}[t]{0.352\textwidth}
        \centering
        \includegraphics[width=\textwidth]{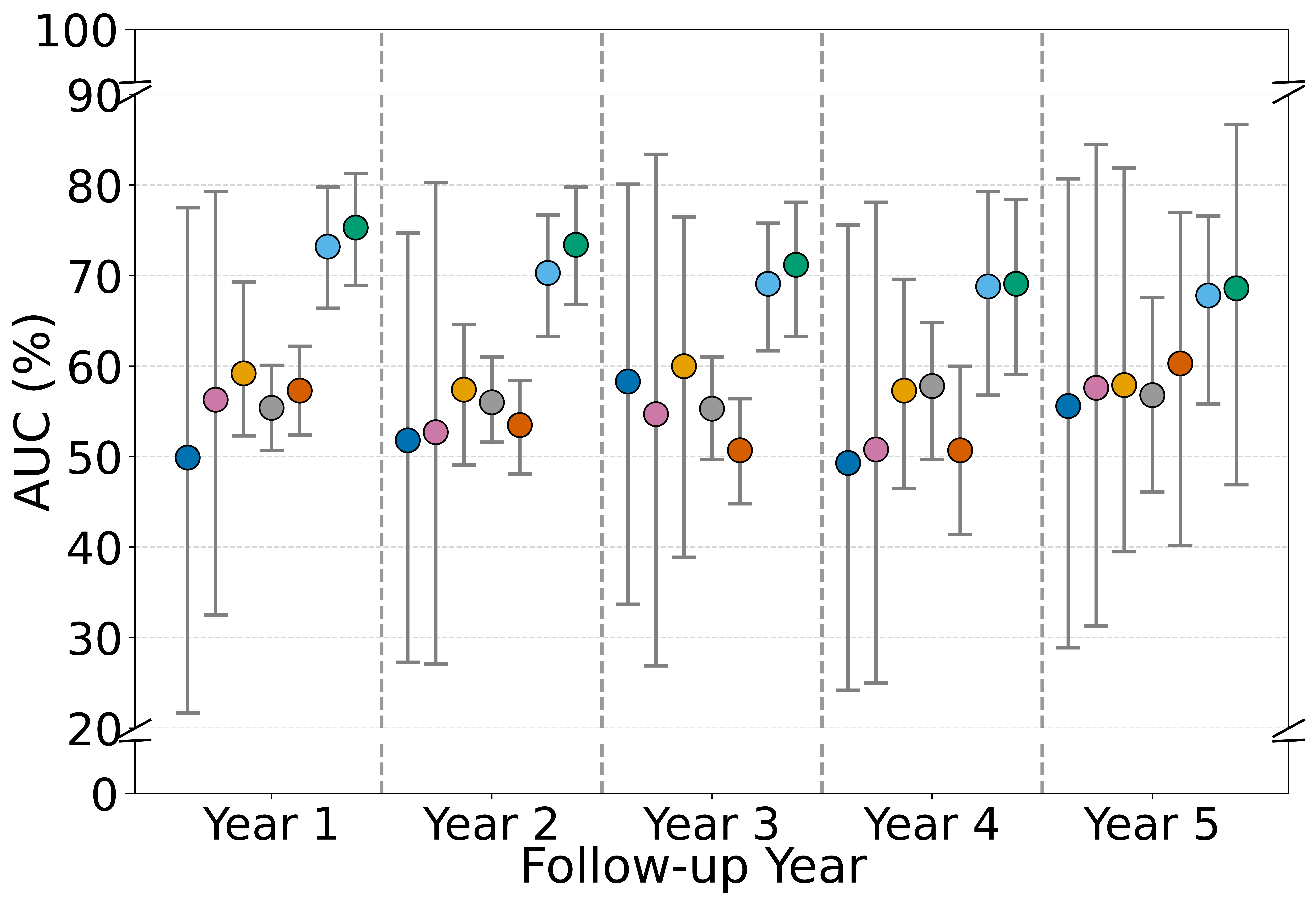}
    \end{minipage}

    \includegraphics[width=0.8\textwidth]{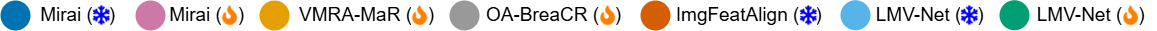}

    \caption{C-index by density category (left; EMBED and CSAW-CC) and by cancer type (middle; EMBED) with 95\% CI, and AUC over follow-up years for invasive cancer (right; EMBED). \textcolor{blue}{\scalebox{0.9}{\faSnowflake}}: frozen encoder,  
\textcolor{orange}{\scalebox{0.9}{\faFire}}: fine-tuned encoder.}
    \label{fig:EMBED_all_combined}
\end{figure}

\noindent \textbf{C-index by density category}
Higher breast density is associated with increased cancer risk~\cite{bodewes2022density} and reduced mammographic sensitivity~\cite{cancerriskdensity}. Fig~\ref{fig:EMBED_all_combined} (left) shows C-index values with 95\% CI across density categories for EMBED and CSAW-CC. Across both datasets, LMV-Net (\textcolor{orange}{\scalebox{0.9}{\faFire}}) achieves the highest C-index across nearly all density groups. Note, some model estimates are not shown for high-density EMBED cases (category D) due to insufficient comparable pairs for reliable estimation. Gains are most pronounced in medium- and high-density groups, highlighting LMV-Net’s robustness in challenging screening scenarios.

\noindent \textbf{Performance across cancer types}
We also perform the first systematic comparison of recent state-of-the-art breast cancer risk prediction methods across different cancer types on the more diverse EMBED dataset. As shown in Fig.~\ref{fig:EMBED_all_combined} (middle), LMV-Net outperforms SOTA methods in C-index for both types, with the fine-tuned encoder achieving the best results and competing methods exhibiting greater uncertainty. Fig.~\ref{fig:EMBED_all_combined} (right) shows AUC over follow-up years for invasive cancer, the more aggressive and clinically relevant type. LMV-Net achieves higher and more stable AUCs, whereas other methods show lower performance and greater uncertainty. These findings demonstrate that integrating multiview and longitudinal information in alignment with clinical workflow yields more robust and reliable breast cancer risk prediction, particularly for invasive disease.
 
\begin{figure}[tb]
    \centering
    \begin{minipage}[b]{0.82\textwidth}
        \centering
        \includegraphics[width=\textwidth]{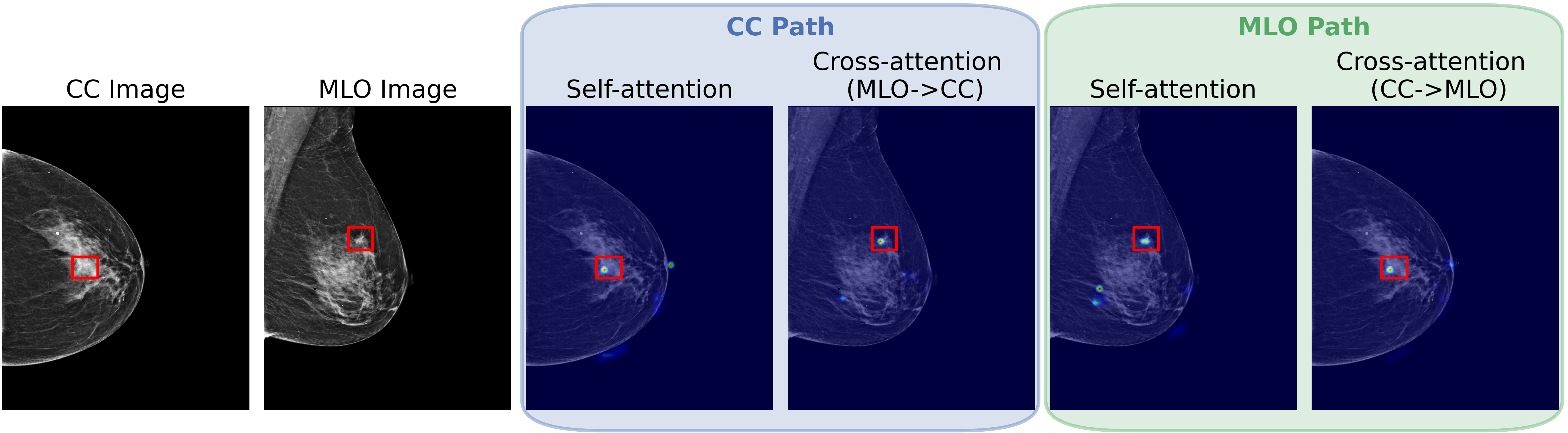}
    \end{minipage}%
    \hspace{0.005\textwidth}
    \begin{minipage}[b]{0.15\textwidth}
        \centering
        \includegraphics[width=\textwidth]{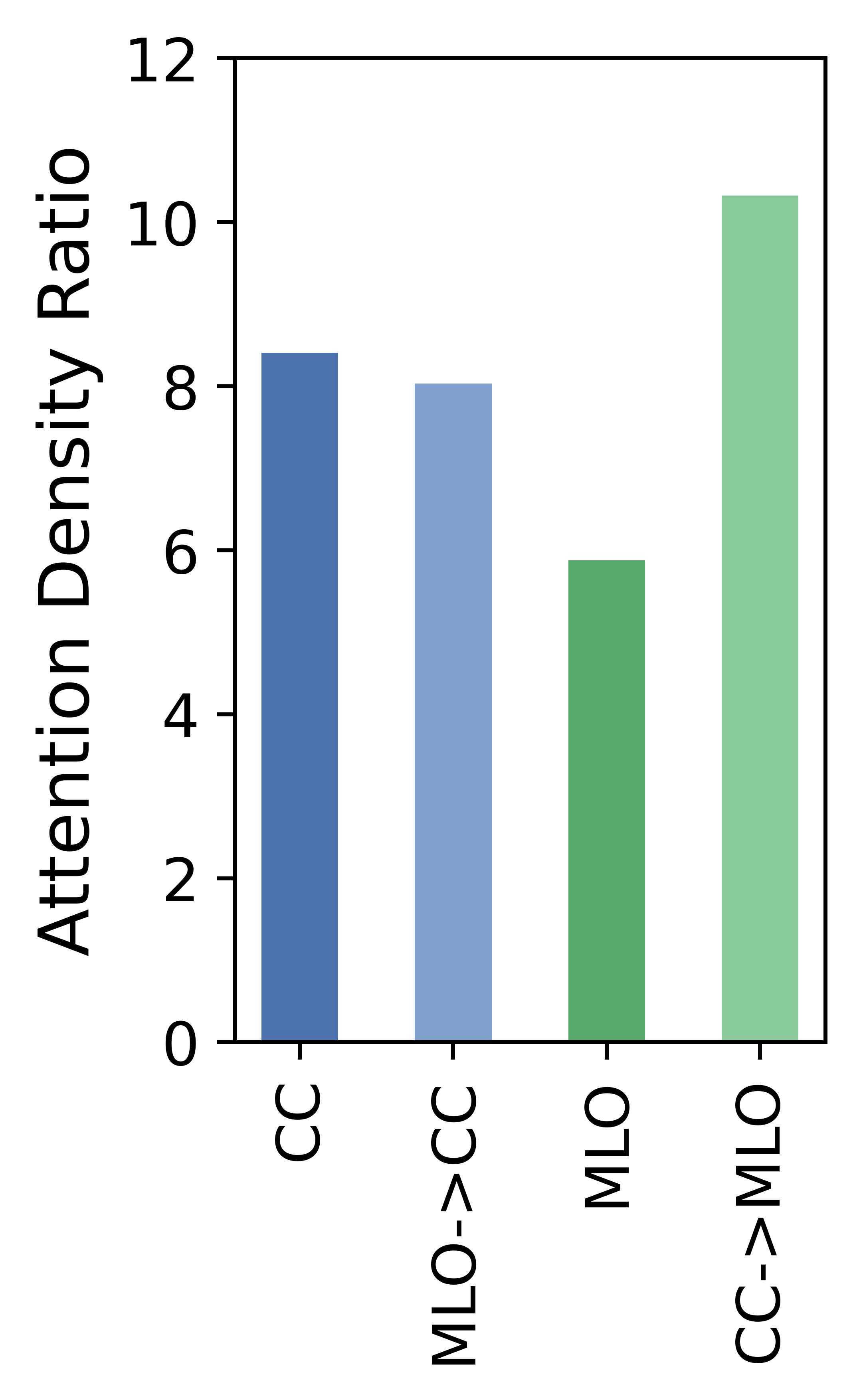}
    \end{minipage}
    \caption{Left: Attention maps showing LMV-Net focus on the tumor (red box). Right: Attention density in tumor bounding box relative to the whole breast.}
    \label{fig:attentionmaps_combined}
\end{figure}

\noindent \textbf{Attention map analysis}
Fig.~\ref{fig:attentionmaps_combined} (left) visualizes the self- and cross-attention maps for a representative patient. LMV-Net assigns elevated attention to the tumor region (red bounding box) in both CC and MLO views. Self-attention primarily focuses on the lesion and its surrounding area within each view, capturing intra-view dependencies. Cross-attention, integrates information from the complementary view, where we see the model attending primarily to the tumor tissue and/or suspicious regions in the other view. 
We further compute the ratio of mean attention density within the tumor bounding box to that within the entire breast region, aggregated across the 12 test-set patients with tumor annotations. (Fig.~\ref{fig:attentionmaps_combined}, right). On average, attention density within the tumor region is 8–10× higher than in the surrounding breast tissue. This supports that the proposed Dual Stream Attention block concentrates attention on anatomically corresponding tumor regions.
 
\noindent \textbf{Ablation study}
We assess key architectural components through ablation studies:
(i) \textit{Repeating current features:} both temporal alignment (Step~2) and subtraction (Step~3) are removed, and Step~4 concatenates the current features three times;
(ii) \textit{Concatenating with zeros:} temporal alignment and subtraction are removed as above, but Step~4 concatenates the current features with two zero tensors;
(iii) \textit{Single-view modeling:} one branch of the model (CC or MLO) is removed, and the cross-attention block in the remaining branch is replaced with an additional self-attention block. Only a single view is processed, effectively disabling multi-view input while preserving the same number of attention layers.
(iv) \textit{Duplicated-view modeling:} the same view is fed into both branches;
(v) \textit{Dual-stream attention removed:} the dual-stream attention block is removed;
(vi) \textit{Multi-view learning:} using only $Y_{\text{multi-view}}$; and
(vii) \textit{Encoder fine-tuning:} freezing the pretrained Mirai backbone.
Table~\ref{tab:ablation_study_final} shows that removing longitudinal information, multi-view modeling, attention, or encoder fine-tuning consistently reduces C-index and AUC. The full model achieves the best performance on both EMBED and CSAW-CC, confirming the value of each component.

\begin{table*}[tb]
\centering
\scriptsize
\caption{Ablation study for 1–5 year risk: mean C-index and AUC on EMBED and CSAW-CC. MV: Multi-view, DV: Different views in both branches, FT: Encoder fine-tuning, DSA: Dual Stream Attention.}
\begin{adjustbox}{max width=\textwidth}
\begin{tabular}{|l|l|c|c|c|c|c|c|c|c|c|c|c|c|}
\hline
\multirow{2}{*}{} &
\multirow{2}{*}{ Variant} &
\multirow{2}{*}{Prior} &
\multirow{2}{*}{MV} &
\multirow{2}{*}{DV} &
\multirow{2}{*}{FT} &
\multirow{2}{*}{DSA} &
\multirow{2}{*}{\makecell{Y$_{\text{MLO}}$\\Y$_{\text{CC}}$}} &
\multirow{2}{*}{C-index (\%) $\uparrow$} &
\multicolumn{5}{c|}{Follow-up year AUC (\%) $\uparrow$} \\

\cline{10-14}
&&&&&&&&& 1-Y & 2-Y & 3-Y & 4-Y & 5-Y \\ 
\hline
\multirow{8}{*}{\rotatebox{90}{\textbf{EMBED}}}
& (i) No prior (current repeated)        & \xmark & \cmark & \cmark & \cmark & \cmark & \cmark & 77.8 & 80.3 & 78.1 & 76.9 & 76.3 & 73.6 \\
& (ii) No prior (zero-padded)   & \xmark & \cmark & \cmark & \cmark & \cmark & \cmark & 77.5  & 80.2 & 78.3 & 76.4 & 76.2 & 73.8 \\
& (iii) Only one view          & \cmark & \xmark & \cmark & \cmark & \cmark & \cmark & 70.8 & 72.2 & 70.8 & 68.9 & 65.9 & 61.9 \\
& (iv) Same view both branches& \cmark & \cmark & \xmark & \cmark & \cmark & \cmark & 77.4  & 78.8 & 78.1 & 77.7 & 76.2 & 74.7 \\
& (v) Without DSA     & \cmark & \cmark & \cmark & \cmark & \xmark & \cmark & 72.8 & 73.9 & 73.7 & 72.8 & 70.3 & 65.8 \\
& (vi) Use only Y$_\text{multi-view}$ & \cmark & \cmark & \cmark & \cmark & \cmark & \xmark & 80.1 & 83.0 & 80.4 & 79.4 & 78.8 & 76.5 \\
& (vii) No finetuning           & \cmark & \cmark & \cmark & \xmark & \cmark & \cmark & 78.3 & 80.8 & 78.3 & 77.5 & 76.6 & 75.1 \\
& \textbf{Ours} & \cmark & \cmark & \cmark & \cmark & \cmark & \cmark & \textbf{81.4} & \textbf{83.9} & \textbf{82.1} & \textbf{80.8} & \textbf{80.0} & \textbf{77.2} \\
\hline
\multirow{8}{*}{\rotatebox{90}{\textbf{CSAW-CC}}} 
& (i) No prior (current repeated)        & \xmark & \cmark & \cmark & \cmark & \cmark & \cmark & 73.0  & 75.3 & 73.8 & 73.2 & 72.3 & 73.2 \\
& (ii) No prior (zero-padded)   & \xmark & \cmark & \cmark & \cmark & \cmark & \cmark & 72.2  & 75.2 & 73.0 & 72.6 & 71.9 & 72.4 \\
& (iii) Only one view          & \cmark & \xmark & \cmark & \cmark & \cmark & \cmark & 73.3  & 73.8 & 73.1 & 73.5 & 73.1 & 73.9 \\
& (iv) Same view both branches& \cmark & \cmark & \xmark & \cmark & \cmark & \cmark & 72.2  & 75.2 & 72.6 & 72.5 & 71.1 & 73.0 \\
& (v) Without DSA     & \cmark & \cmark & \cmark & \cmark & \xmark & \cmark & 69.5  & 70.5 & 70.0 & 70.6 & 69.3 & 71.4 \\
& (vi) Use only Y$_\text{multi-view}$ & \cmark & \cmark & \cmark & \cmark & \cmark & \xmark & 73.3 & 77.6 & 73.3 & 73.4 & 72.6 & 74.4 \\
& (vii) No finetuning           & \cmark & \cmark & \cmark & \xmark & \cmark & \cmark & 73.4  & 77.5 & 73.5 & 74.0 & 73.1 & \textbf{75.9} \\
& \textbf{Ours} & \cmark & \cmark & \cmark & \cmark & \cmark & \cmark & \textbf{74.4} & \textbf{78.7} & \textbf{74.1} & \textbf{74.3} & \textbf{73.5} & 75.0 \\
\hline
\end{tabular}
\end{adjustbox}
\label{tab:ablation_study_final}
\end{table*}

\section{Conclusion and Outlook}
In this study, we introduced LMV-Net, which mimics the radiologist's workflow by jointly modeling longitudinal and multi-view information through explicit longitudinal alignment. Experiments on two large publicly available datasets demonstrate statistically significant improvements over SOTA methods, particularly in challenging scenarios such as patients with higher breast density. Clinically, this approach could improve early detection and personalized screening through more accurate risk assessment. While our results demonstrate the value of joint longitudinal multi-view modeling, further investigation is warranted. Incorporating multiple prior examinations may improve performance by providing richer temporal context. In addition, our framework relies on explicit feature-space alignment to enable accurate longitudinal modeling such that advances in robust explicit alignment methods could further enhance its performance. Future work will also focus on more comprehensive evaluation across additional clinically relevant metrics, including precision–recall AUC (PR-AUC), particularly given the class imbalance characteristic of breast cancer risk prediction.

%
%
%
%
\bibliographystyle{splncs04}
\bibliography{Paper-2399}

\end{document}